\documentclass[lettersize,journal]{IEEEtran}

\usepackage{amsmath,amsfonts}
\usepackage{multirow}
\usepackage{multicol}

\DeclareMathOperator*{\argmax}{arg\,max}
\usepackage{gregmath}
\usepackage{algorithm}
\usepackage{algcompatible}
\renewcommand{\algorithmiccomment}[1]{\bgroup\hfill//~#1\egroup}
\usepackage{nicefrac}
\usepackage{array}
\usepackage[caption=false,font=normalsize,labelfont=sf,textfont=sf]{subfig}
\usepackage{textcomp}
\usepackage{stfloats}
\usepackage{url}
\usepackage{verbatim}
\usepackage{graphicx}
\usepackage{cite}
\usepackage{booktabs}
\newcommand{\mathbbm}[1]{\text{\usefont{U}{bbm}{m}{n}#1}}
\usepackage[table,x11names]{xcolor}
\begin{document}

\title{A Maximum Log-Likelihood Method for Imbalanced Few-Shot Learning Tasks}
\author{
    Samuel~Hess and~Gregory~Ditzler,~\IEEEmembership{Senior~Member,~IEEE}
    \thanks{
        S. Hess is with the Department of Electrical and Computer Engineering, University of Arizona, Tucson, AZ, 85721.  
        E-mail: \texttt{shess@arizona.edu} 
    }
    \thanks{
        G. Ditzler is with the Department of Electrical and Computer Engineering, Rowan University, Glassboro, NJ, 08028.
        E-mail: \texttt{ditzler@rowan.edu}
    }
    \thanks{This work was supported by the National Science Foundation (NSF) CAREER \#1943552.}
}

\maketitle

\begin{abstract}
Few-shot learning is a rapidly evolving area of research in machine learning where the goal is to classify unlabeled data with only one or ``a few'' labeled exemplary samples. Neural networks are typically trained to minimize a distance metric between labeled exemplary samples and a query set. Early few-shot approaches use an episodic training process to sub-sample the training data into few-shot batches. This training process matches the sub-sampling done on evaluation. Recently, conventional supervised training coupled with a cosine distance has achieved superior performance for few-shot. Despite the diversity of few-shot approaches over the past decade, most methods still rely on the cosine or Euclidean distance layer between the latent features of the trained network. In this work, we investigate the distributions of trained few-shot features and demonstrate that they can be roughly approximated as exponential distributions. Under this assumption of an exponential distribution, we propose a new maximum log-likelihood metric for few-shot architectures. We demonstrate that the proposed metric achieves superior performance accuracy w.r.t. conventional similarity metrics (e.g., cosine, Euclidean, etc.), and achieve state-of-the-art inductive few-shot performance. Further, additional gains can be achieved by carefully combining multiple metrics and neither of our methods require post-processing feature transformations, which are common to many algorithms. Finally, we demonstrate a novel iterative algorithm designed around our maximum log-likelihood approach that achieves state-of-the-art transductive few-shot performance when the evaluation data is imbalanced. We have made our code publicly available at \url{https://github.com/samuelhess/MLL_FSL/}.
\end{abstract}


\section{Introduction}
\IEEEPARstart{F}{ew}-shot neural networks are rapidly emerging as a promising technique to address challenges in conventional deep neural networks (DNNs). Specifically, conventional DNNs have excelled in many classification tasks where large volumes of training data are available for each class \cite{he2016deep, silver2016mastering, oord2016wavenet}. Unfortunately, classification on unseen classes (i.e., classes unobserved at training) causes a dilemma since the output nodes of the DNN are directly trained to a specific class. Many few-shot approaches circumvent this dilemma by using training data to learn a distance metric at the network's multi-dimensional feature layer output (often the flattened layer prior to the classification layer). Ideally, in this multi-dimensional space, samples within the same class are clustered closely into groupings, and samples of different classes are further apart. During evaluation, the network can be queried for a distance between an unknown sample to a set of exemplary known samples from each class. The inferred class for the unknown sample is the class with the minimum distance. Note this type of evaluation within the few-shot learning community is essentially an application of a nearest neighbor classifier. 

\begin{figure}
\centering 
\includegraphics[width=0.48\textwidth]{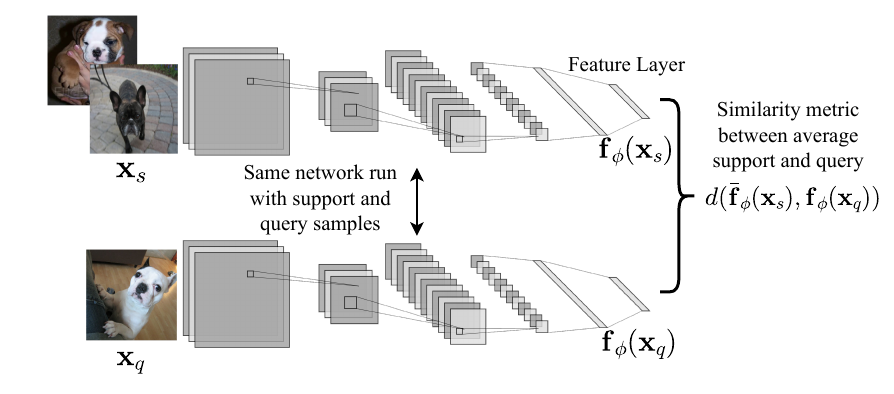}
\caption{Diagram of conventional few-shot architecture. One branch of the architecture ingests exemplary samples of a class, the other branch ingest the query sample under test. A similarity metric, conventionally Euclidean or cosine, is computed between the high dimensional vectors at the final feature layer.}
\label{fig:few-shot_diagram}
\vspace{-1em}
\end{figure}

Many early DNN approaches to few-shot learning use a dual architecture with identical DNN branches that embed nonlinear features of a sample. As illustrated in Figure \ref{fig:few-shot_diagram}, one branch of the architecture ingests a sample from a known class ($\mathbf{x}_{s}$), and the other branch ingests the query sample under test ($\mathbf{x}_{q}$). The output is a comparator (Euclidean or cosine distance metric) between the last feature layer of the two branches. Several modifications have evolved from the literature that improves state-of-the-art performance with novel DNN feature embedding backbones, architectures, training paradigms, and data conditioning. Additionally, it is now commonplace to evaluate both inductive and transductive performance. Inductive evaluation reports statistical performance w.r.t. an arbitrary unknown sample, whereas transductive evaluation reports statistical performance against a {\em{batch}} of unknown samples. In the latter case, the batch of samples provides additional information that can be inferred as a group. Therefore, transductive methods often achieve marginal gains in performance over inductive methods.

Despite the novel designs for few-shot network architectures, Chen et al. showed that using common DNN backbones, training paradigms, and data assumptions results in similar performance \cite{chen2019closer}. Hence, the result of a ``fair'' comparison between the different approaches resulted in insignificant differences in performance. Likewise, Veilleux et al. showed that transductive performance is similar when considering imbalanced data,  which is common to many real-world applications \cite{veilleux2021realistic}. Additionally, many few-shot approaches still rely on a cosine or Euclidean metric (or some close variant) to train the neural network. For example, in very recent work, PT-MAP claimed state-of-the-art transductive performance \cite{hu2021leveraging}. This performance gain is achieved by pretraining a network using supervised learning and then fine-tuning the feature layer(s) to maximize the cosine similarity between samples of the same class. During evaluation, PT-MAP minimizes the Euclidean distance between the distributions of unknown and known samples using an iterative Sinkhorn method \cite{cuturi2013sinkhorn}. The authors' observed that the original features are not Gaussian, and achieve significant performance gain via post-processing steps (i.e., power transform, inductive normalization, and transductive normalization) used to make the features more Guassian-like.

In this work, we investigate the feature distributions and claim that, while the features are not Gaussian distributed, they can be better approximated as exponential distributions. Further, in making this fast approximation, we empirically observe that the distribution of parameters is dependent on the class. This observation allows us to apply a maximum log-likelihood approach as a metric between class samples. We demonstrate that our novel metric (without the need for feature post-processing) consistently outperforms the Euclidean metric, and achieves new state-of-the-art inductive performance on miniImageNet and TieredImageNet datasets. Additional gains can be achieved by combining the Euclidean and cosine with our metric, given that careful statistical considerations are made for the correlations between them. Finally, we demonstrate that our method can provide an estimated posterior probability. This posterior can be used to weight and combine batches of unknown samples into an iterative method for stronger predictions. The iterative method is robust to data imbalances and achieves new state-of-the-art performance in imbalanced transductive few-shot tasks.

\section{Related Work}
Few-shot learning can be broadly partitioned into two taxonomies: inductive and transductive methods. Inductive methods use only the labeled exemplary samples to infer each unlabeled samples' class during testing. Transductive methods achieve superior performance over inductive methods by using the batch of unlabeled samples to help infer all the unlabeled data together. In both cases, the evaluation of the model is performed on a $N$-way, $K$-shot task. The transductive setting additionally assumes that the evaluation has a set of $\mathcal{Q}_{u}$ unlabeled query samples. In the balanced setting, $\mathcal{Q}_{u}$ contains the same number of unlabeled samples from for each of the $N$ classes. In the imbalanced setting, $\mathcal{Q}_{u}$ has a different number of samples for each class. The following subsections formally define both inductive and transductive evaluation and summarizes the related work surrounding each approach.

\subsection{Inductive Few-Shot}
In an $N$-way, $K$-shot task, a model is provided a support set, $\mathcal{S}$, that has $K$  $D$-dimensional samples $\mathbf{x}_i \in \mathbb{R}^D$ with corresponding labels ${y}_i \in \mathcal{Y}$ from $N$ different classes (i.e., $|\mathcal{Y}| = N$). Formally, we can represent $\mathcal{S}$ as a union of the sets from the individual classes (i.e., $\mathcal{S} := \mathcal{S}_1 \cup \cdots \cup \mathcal{S}_N$), 
where $\mathcal{S}_c=\{(\mathbf{x}_{s1}, {y}_{s1}=c), \ldots, (\mathbf{x}_{sK}, {y}_{sK}=c)\}$) and $c\in\mathcal{Y}$. 
Then, given a new (never-before-seen) query sample, $\mathbf{x}_{q}$ (with unknown true label ${y}_{q}$) the task's objective is to estimate the label (i.e., $\widehat{y}_{q}$) with arbitrarily high accuracy. Naturally, the task becomes significantly more challenging as the number of exemplary samples in the support set ($K$) decreases, or the number of classes ($N$) increases.

Few-shot has been pursued by machine learning researchers for decades \cite{fink2004object, fei2006one}; however, the field has rapidly expanded over the past few years due to the development of standard benchmark datasets, and key advancements in neural networks and computing. The first prominent DNN approach for few-shot was Siamese Networks developed by Koch et al. \cite{koch2015siamese}. As the name implies, Siamese Networks create an architecture of two identical networks (i.e., networks with tied weights) as a comparator between two samples. The networks take an input from a known class $(\mathbf{x}_{s1},{y}_{s1}=c)$ then  pass it through one branch of the network ${\phi}$ to obtain a feature representation $\mathbf{f}_{\phi}(\mathbf{x}_{s1})$. Similarly, a query sample $\mathbf{x}_{q}$ passes through an identical network to obtain a feature representation $\mathbf{f}_{\phi}(\mathbf{x}_{q})$. Note that the vector $\mathbf{f}_{\phi}(\mathbf{x}_{q})$ is the from a (latent) feature layer of a DNN with parameters $\phi$. 
The Euclidean distance,  $\left\Vert \mathbf{f}_{\phi}(\mathbf{x}_{s1}) - \mathbf{f}_{\phi}(\mathbf{x}_{q}) \right\Vert_2^2$, is a dissimilarity metric that allows the Siamese Network to infer if $\mathbf{x}_{q}$ belongs to the same class as $\mathbf{x}_{s1}$. The distance is then passed through an activation function and the binary cross entropy loss is used to train the network. 
The activation function after the Euclidean distance is simply zero or one if the samples are from the same or different classes, respectively.
Siamese Networks were initially evaluated on the Omniglot dataset, which contains 20 handwritten characters from 50 different alphabets \cite{lake2011one}. The entire dataset totals 32,460 characters (20 samples of 1,623 characters), and the author's of Siamese Networks partitioned the characters into disjoint class sets for training, validation, and testing. The results from Siamese Networks were the first to demonstrate neural network performance on the one-shot Omniglot task.

Shortly following the development of Siamese Networks, Vinyals et al. \cite{vinyals2016matching} proposed episodic training for Matching Networks. Episodic training considers  subsets of the training data that are repeatedly (and randomly with replacement) selected to ``match'' the few-shot evaluation. That is, given an $N$-way, $K$-shot evaluation task, Matching Networks sub-sample the training data with a random set of $N$-way, $K$-shot, support samples and $M$ query samples for each training episode. 
Matching networks use tied weights (similar to Siamese networks), they compute a cosine distance metric between a query sample and each support sample, and they use an attention mechanism to select the most likely class. 
Vinyals et al.  also introduced a more challenging few-shot benchmark of real images, namely miniImageNet, which is now the {\em de facto} benchmark for few-shot performance \cite{vinyals2016matching}. Additionally, most metric-based approaches include the episodic training scheme; however, recent approaches have outperformed episodic training methods using a combination of standard supervised training and then fine-tuning to transfer to the few-shot task \cite{hu2021leveraging, mangla2020charting, wang2019simpleshot, tian2020rethinking}.

Prototypical Networks are another prominent few-shot baseline that follows the episodic training paradigm from Matching Networks. 
However, Prototypical networks compute the Euclidean distance at the feature layer (instead of the cosine distance used in Matching Networks), and the distance is computed between a query sample and the average (i.e. prototype) support sample feature \cite{snell2017prototypical}. 
In contrast to the metric learning approach of Matching and Prototypical Networks, Model-Agnostic Meta-Learning (MAML) \cite{finn2017model-SAMBIB}, and Meta-Learner LSTM \cite{ravi2016optimization} are prominent works that learn to fine-tune to new tasks rapidly. MAML and the many derivative works \cite{yoon2018bayesian, finn2018probabilistic} use batch normalization. Unfortunately, Bronskill et al. correctly notes that batch normalization with the fine-tuning stage, benefits from the statistical information across the batch of query samples \cite{bronskill2020tasknorm}. Therefore, MAML (with batch normalization) is technically a transductive, rather than inductive approach.

There are many few-shot methods that are more complicated than Prototypical Networks, Matching Networks, and MAML. Such approaches include metric architectures \cite{ye2018deep, scott2018adapted}, generative architectures \cite{wang2018low, antoniou2017data}, fine tuning architectures \cite{rusu2018meta, jiang2018learning}, or some combination of these approaches; however, Chen et al. showed many of the few-shot methods have comparable results when using the same neural network backbone and training paradigm \cite{chen2019closer}. As architectures continue to become more complicated, consistency in their inductive evaluation becomes more challenging.
\vspace{-1em}

\subsection{Transductive Few-Shot}
\label{sec:transductive_related_works}
It is common practice to evaluate inductive performance on a set of query batches. A query set, $\mathcal{Q}$ is defined similar to a support set where $\mathcal{Q} := \mathcal{Q}_1 \cup \cdots \cup \mathcal{Q}_N$, and $\mathcal{Q}_c=\{(\mathbf{x}_{q1},{y}_{q1}=c),\ldots,(\mathbf{x}_{qM_{c}},{y}_{qM_{c}}=c)\}$  is a balanced dataset of $M_{c}$ never-before-seen query samples for each of the $N$ classes of the support set. The inductive objective provides an estimated label for each of the respective query samples in the batch, independent of any other query sample. In contrast, transductive few-shot performance takes into account the unlabeled query set as a whole. That is, given the entire labeled support set $\mathcal{S}$, and the entire unlabeled query set $\mathcal{Q}_{u} := \mathcal{Q}_{u1} \cup \cdots \cup \mathcal{Q}_{uN} =\{\mathbf{x}_{q1},\ldots,\mathbf{x}_{qM}\}$, the transductive few-shot task estimates the entire set of labels $\{\widehat{y}_{q1},\ldots,\widehat{y}_{qM}\}$. Unlike the inductive task, the outcome of each estimate, $\widehat{y}_{qi}$ is dependent on the entire query set. The entire unlabeled query set has additional information, via the other unlabeled query samples, that give transductive few-shot approaches an edge over inductive approaches w.r.t classification accuracy.

The first prominent DNN work to address transductive few-shot was Liu et al.'s  Transductive Propagation Network (TPN) \cite{liu2018learning}. TPN passes support and query samples through a network to produce feature embeddings, $\mathbf{f}_{\phi}(\mathbf{x}_{i})$, and then these embeddings are passed through another network to learn an example-wise scaling parameter $\mathbf{g}_{\phi}(\mathbf{f}_{\phi}(\mathbf{x}_{i}))$. A weighted graph is computed by sampling every pairwise combination of the support and query set as
\[
\mathbf{W}_{ij} = \exp \left(-\frac{1}{2} \left\Vert \frac{\mathbf{f}_{\phi}(\mathbf{x}_{i})}{\mathbf{g}_{\phi}(\mathbf{f}_{\phi}(\mathbf{x}_{i}))} - \frac{\mathbf{f}_{\phi}(\mathbf{x}_{j})}{\mathbf{g}_{\phi}(\mathbf{f}_{\phi}(\mathbf{x}_{j}))} \right\Vert_2^2 \right)
\]
where the $k$-nearest neighbors are selected from each row and the division of the vectors is performed example-wise. Labels for unknown query samples are determined from support samples via label propagation \cite{zhou2003learning}.

There  has been a large breadth of research on transductive few-shot since TPN, including variants of graph neural networks \cite{huang2020ptn, yang2020dpgn}, attention-based mechanisms \cite{hou2019cross, guo2020attentive}, and meta-learning gradient methods \cite{hu2020empirical}. Several post-hoc methods (i.e., methods that operate on features extracted from existing network architectures) have also emerged in transductive few-shot literature, including transductive fine-tuning \cite{dhillon2019baseline}, Laplacian regularization \cite{ziko2020laplacian}, and group inference methods \cite{liu2020prototype,wang2020instance,hu2021leveraging}. To the best of our knowledge, \cite{hu2021leveraging} have notably achieved the best 1-shot transductive performance on benchmark datasets, achieving 82.9\% and 85.7\% classification accuracy on miniImageNet and tieredImageNet, respectively. Their approach, namely PT-MAP, is a post-hoc method that applies a power transform to produce features from the few-shot network to look like a Gaussian feature. PT-MAP uses a self-supervised manifold mixup architecture that is trained using Mangla et al.'s  approach \cite{mangla2020charting}. 
After the set of all support and query samples go through a power transform, statistics are computed and a normalization is applied to the entire set. The normalization across the entire set is the first transductive aspect of their approach because it requires information about query samples. 
The features also go through QR decomposition to reduce the dimensionality. The dimensionality reduction step was shown to reduce the computational complexity of PT-MAP without a noticeable degradation in performance. Finally, they achieve the majority of their performance gain by iteratively updating the class centers using a Sinkhorn mapping process. The transducitve multi-class Sinkhorn mapping uses the information from the full set of modified support and query features as well as the fact that the evaluation task is balanced (i.e., there are exactly the same amount of support and query samples for each class).

Veilleux et al. recently showed that many transductive few-shot approaches, such as PT-MAP, assume that the marginal probability of the classes is fixed and uniform, which is untrue for many real-world applications \cite{veilleux2021realistic}. Further, Veilleux et al. modify the proportions of classes to be imbalanced by sampling from a Dirichlet distribution and demonstrate that several few-shot model's performance are significantly degraded. They also proposed a new formulation of the neural network loss, called $\alpha$-TIM, which computes the generalized mutual information between support and query samples. In contrast to the marginal cross entropy loss used by many prior neural network classifiers, the mutual information includes the computation of both the marginal and conditional entropy \cite{veilleux2021realistic, boudiaf2020information}. $\alpha$-TIM demonstrated state-of-the-art accuracy on transductive imbalanced classification.

\section{The Maximum Log-Likelihood Approach}
\subsection{Motivation}
\label{sec:Motivation}
The feature layer of a neural network can be represented by a nonlinear embedding function $\fbf(\cdot): \mathbb{R}^D \mapsto \mathbb{R}^I$ that transforms a $D$-dimensional input sample $\mathbf{x} \in \mathbb{R}^D$ (e.g., an image) into a $I$-dimensional feature representation at the hidden layer of a network. 
The core assumption of most machine learning models is that the inputs $\mathbf{x}$ are sampled from fixed -- albeit unknown -- probability distribution \cite{ditzler2015learning}, where these random variables are passed through multiply-and-accumulate operations followed by a nonlinear function (e.g., a ReLU). Each element $f_{i}(\mathbf{x})$ of the neural network embedding $\fbf(\mathbf{x})$ is a nonlinear transformation of random variable $\mathbf{x}$.
In this work, we empirically investigate the distribution of these nonlinear features for few-shot networks and show that an exponential distribution conveniently provides a fast approximation for the $i$th feature node (i.e., $f_{i}(\cdot)$) when the last nonlinear activation is ReLU.

Formally, given a set of samples from the same class, $a$, we define a set  $\mathcal{S}_{a}=\{(\mathbf{x}_{1},{y}_{1}=a), \ldots, (\mathbf{x}_{L},{y}_{L_{a}}=a)\}$. 
The samples $\mathbf{x}_{l}$ are passed through the network to obtain the feature representations $\mathbf{f}(\mathbf{x}_{l})$. Accurately modeling the multivariate hidden feature representation's distribution is non-trivial; however, the distribution of a single random variable can be roughly approximated by using a histogram to estimate $\widehat{p}_{i}(z)$, which is given by: 
\begin{equation}
    \widehat{p}_{i}(z) = \frac{1}{LB} \sum_{l = 1}^{L} \mathbbm{1}_{ \left[ z-\frac{B}{2} \leq f_{i}({\xbf}_{l}) < z+\frac{B}{2}\right]}
\end{equation}
where $z$ is the discrete bin center that increments in steps of the bin size ($B$), and $\mathbbm{1}_{[\cdot]}$ is the indicator function.

\begin{figure*}
     \centering
     \subfloat[Euclidean Features]{\includegraphics[width=.45\textwidth]{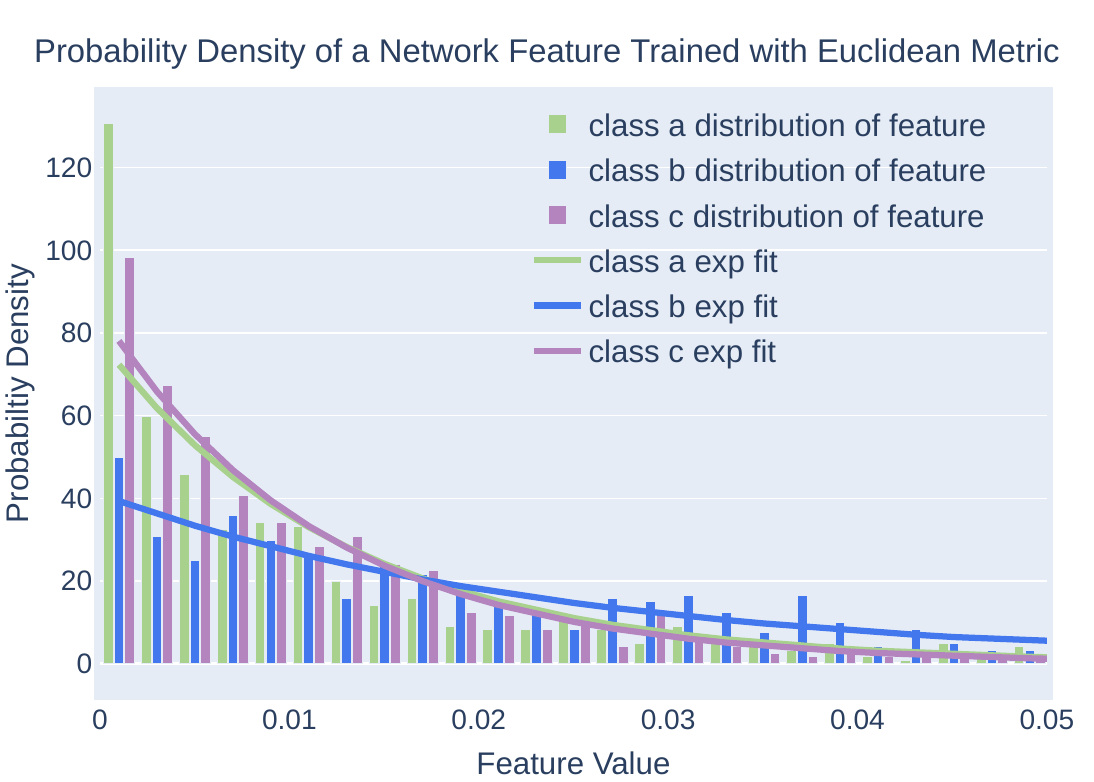}\label{fig:FEAT_Euclidean_Features}} 
     \subfloat[Cosine Features]{\includegraphics[width=.45\textwidth]{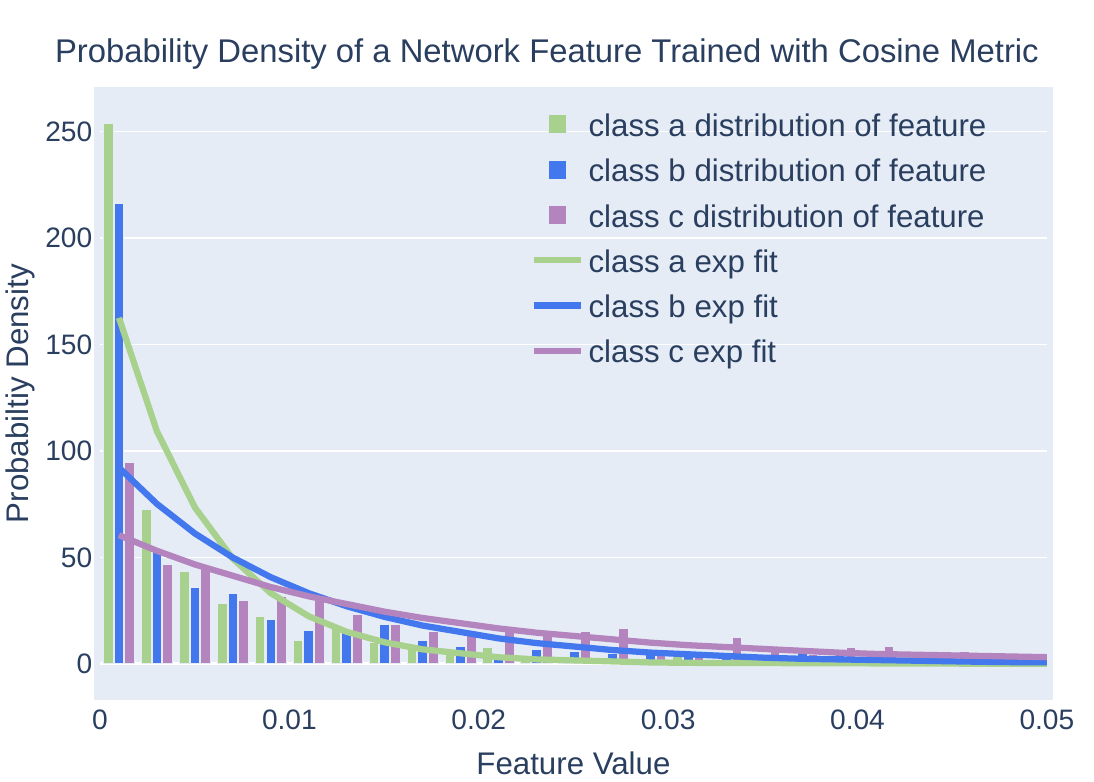}\label{fig:FEAT_Cosine_Features}} 
         \caption{Visualization of the distributions from the $i$th feature of a network trained with the prototypical paradigm in \cite{ye2020few}. The histograms are the empirical distributions from the same $i$th feature of a random class of miniImageNet validation data. The corresponding lines represent the approximate maximum likelihood estimation of the distribution w.r.t to an exponential fit. Different classes are observed to have different parameters for the exponential distribution.}
        \label{fig:feature_examples}
\end{figure*}

Consider the histrogram shown in Figure \ref{fig:feature_examples}. 
This figure shows the empirical probability density distribution (PDF) of a single feature ($f_{i}(\cdot)$) at the last hidden layer of a prototypical network using ReLU activation trained on miniImageNet. 
Note this figure only shows the distribution of the validation dataset. 
The architecture used to produce this plot is a prototypical network that was trained using the paradigm developed by Ye et al. \cite{ye2020few}. Two prototypical networks were trained, one with a Euclidean metric (see Figure \ref{fig:FEAT_Euclidean_Features}) and one with a cosine metric (see Figure \ref{fig:FEAT_Cosine_Features}).  
Regardless of the distance metric, we observe that the feature density is heavily weighted towards zero and decreases as a function of the feature magnitude. This behavior is similar to an exponential distribution (i.e., $\widehat{p}_{i}(z) \approx {\lambda}_{a}e^{-{\lambda}_{a}f_{i}(\mathbf{x}_{l})}$). 
Using maximum likelihood estimation, the parameter of the distribution can be estimated as  
\begin{align}
    {\lambda_{a}} = \frac{L}{\sum_{l = 1}^{L}f_{i}({\xbf}_{l})}.
\end{align}

It is further observed that the single parameter of the distribution varies w.r.t class (i.e., the exponential parameter $\lambda$ would be different for each class). 
In general, we observe that ${\lambda}_{a} \neq {\lambda}_{b}$ for two distinct classes $a$ and $b$ as seen from the approximated distributions in Figure \ref{fig:feature_examples}. 

The results shown in Figure \ref{fig:feature_examples} represent the distribution from a single feature of a particular trained network (i.e. $f_{i}(\cdot)$); however, our code repository allows users to further investigate all features. We have found our observations generally hold across the elements of the feature vector and several few-shot architectures that use different training paradigms and loss functions. It is important to note that the feature data are not strictly exponentially distributed, especially very close to zero where the density is largest; however, by making this assumption of an exponential distribution, a maximum likelihood (MLE) approach can be easily applied to estimate the statistics of the distribution from labeled samples, and subsequently, classify unlabeled samples. In contrast to conventional metrics, our classification metric is motivated by the expected distributions of the feature data. 

\subsection{Definition of the Maximum Log-Likelihood metric for Few-Shot Networks}
\label{sec:MLL_Metric}
In this section, we introduce our proposed approach to estimate the exponential distribution parameter of a feature and use it to classify unlabeled samples. Formally, given a $K$-shot, $N$-way few-shot task, we define a support set for each class $c$ as $\mathcal{S}_{c} = \{(\mathbf{x}_{1}, {y}_{1}=c), \ldots, (\mathbf{x}_{K}, {y}_{K}=c)\}$. The MLE of ${\lambda}_{c,i}$ from the $i$th feature is computed from the known support set as
\begin{equation} \label{eq:param_est}
{\lambda}_{c,i} =  \frac{|\mathcal{S}_c|}{ \sum\limits_{(\xbf_{k},y_{k})\in S_c}f_{i}({\xbf}_{k})}
\end{equation}
Given a query sample $\mathbf{x}_{q}$ with unknown label $y_{q}$, the probability the sample belongs to class $c$ is then given by the exponential probability density function: 
\begin{equation}
\widehat{p}_{i}(\mathbf{x}_{q}) = {\lambda}_{c,i}\,\text{e}^{-{\lambda}_{c,i}\, f_{i}(\mathbf{x}_{q})}
\end{equation}
The class that maximizes this probability (or equivalently maximizes the log probability), is the most likely estimate ($\widehat{y}_{q,i}$) of the class for the $i$th feature. That is
\begin{equation}
\widehat{y}_{q,i} = \argmax_{c\in N} \left[\log({\lambda}_{c,i}) - {\lambda}_{c,i}f_{i}(\mathbf{x}_{q})\right]
\end{equation}
This equation applies only to the prediction using the $i$th feature. 
By na\"ively assuming independence across all the features, the log probabilities can be summed, providing the most likely estimate $\widehat{y}_{q}$ as
\begin{equation} 
\widehat{y}_{q} = \argmax_{c\in N} \left[\sum_{i=1}^I\log({\lambda}_{c,i}) -  \boldsymbol{\lambda}_{c} \cdot \mathbf{f}(\mathbf{x}_{q}) \right] 
\label{eq:MLL}
\end{equation}
where $\boldsymbol{\lambda}_{c}$ represents the vector of estimated parameters for each node in the network.
In practice, it is unlikely that the features are independent; however, similar to the na\"ive Bayes and logistic regression, the independence assumption further simplifies our method and still leads to the state-of-the-art performance as shown in Section \ref{sec:experiments}. Incorporating the feature dependence with a multivariate exponential distribution might yield better results, however, this task is significantly more challenging to compute. Therefore, we leave relaxing the independence assumption to future work. 

Note that the metric inside the argument of equation \eqref{eq:MLL} is differentiable and, therefore, it can be incorporated as a loss in metric-based few-shot neural networks training paradigms. Algorithm \ref{alg:MLL} provides the pseudo-code of our approach with a prototypical network architecture and an episodic training paradigm. The algorithm introduces our metric on lines 7--9. Line 8 adds a hyperparameter (${\lambda}_{max}$) to clip ${\lambda}_{c,i}$. ${\lambda}_{max}$ stabilizes training in early iterations where large $\lambda$ estimates (feature values very close to zero) cause the logarithmic term to extend outside floating point precision. During training we found that ${\lambda}_{max}$ can be set very high. We present results with this parameter set to 100 and is sufficient to obtain stable training results. After training, we perform a parameter sweep between 10 and 100 on the validation data with increments of 10 and use the one value that achieves the best performance across all networks. The optimal clip value on the validation data was 40 and is subsequently used for evaluation.

 \begin{algorithm*}[t] \footnotesize
 \caption{Prototypical Network with Maximum Log Likelihood Metric pseudo-code}\label{alg:MLL}
 \begin{algorithmic}[1]
 \renewcommand{\algorithmicrequire}{\textbf{Input:}}
 \renewcommand{\algorithmicensure}{\textbf{Output:}}
 \REQUIRE{Entire Training Set $Tr=\{(\mathbf{x}_{1},{y}_{1}),...,(\mathbf{x}_{N_{Tr}},{y}_{N_{Tr}})\}$, number of classes per episode $N$, number of support samples per class $K$, number of query samples per episode $M$, clip parameter ${\lambda}_{max}$, initial model $\phi$, and model learning rate $\eta$}
 \ENSURE{updated model $\phi$, network loss $J(\phi)$, and class estimates $\widehat{y}_{q_m}$}
  \FOR {each episode in training data}
  \STATE Create a set, $\mathcal{E}=\{(\mathbf{x}_{1},{y}_{1}),...,(\mathbf{x}_{N_{E}},{y}_{N_{TE}})\}$, of $N$ different classes randomly chosen from $Tr$ without replacement
  \STATE Create disjoint sets, $\mathcal{S}_{c}=\{(\mathbf{x}_{s1},{y}_{s1}),...,(\mathbf{x}_{sK},{y}_{sK})\}$ and $\mathcal{Q}=\{(\mathbf{x}_{q1},{y}_{q1}),...,(\mathbf{x}_{qM},{y}_{qM})\}$, of $K$ support examples and M query samples from $E$ 
   \FOR[Loop through query samples]{$m = 1,\ldots,M$}
   \FOR[Loop through class samples]{$c = 1,\ldots,N$}
   \STATE // Compute the classification scores for every query sample
   \STATE ${\lambda}_{c,i} =  \nicefrac{|S_c|}{ \sum_{(\xbf_{sk},y_{sk})\in
   S_c}f_{i}({\xbf}_{sk})}$ \COMMENT{Equation \eqref{eq:param_est}}
   \STATE ${\lambda}_{c,i} = \min({\lambda}_{c,i},{\lambda}_{max})$
   \STATE $\alpha_{qm}(c)= \sum_{i}\log({\lambda}_{c,i}) -  \boldsymbol{\lambda}_{c} \cdot \mathbf{f}(\mathbf{x}_{qm})$ \COMMENT{Equation \eqref{eq:MLL}}
   \ENDFOR
   \FOR{$c = 1,\ldots,N$}
   \STATE $p_\phi(y=c | \xbf_{qm}) = \frac{e^{\alpha_{qm}(c)}} {\sum_{c'\in N} e^{\alpha_{qm}(c')}}$ \COMMENT{Estimate posterior via softmax}
   \ENDFOR
   \STATE $\widehat{y}_{q_m} = \argmax_{c\in N}[p_\phi(y=c | \xbf_{q_m})]$ \COMMENT{Estimate class of query sample}
  \ENDFOR
  \STATE $J(\phi) = -\frac{1}{NM}\sum_{c' = 1}^{N}\sum_{q = 1}^{M}\log(p_\phi(y=c' | \xbf_{qm}))$ \COMMENT{Compute average loss of episode}
  \STATE $\phi \leftarrow \phi - \eta\nabla_{\phi}J(\phi)$ \COMMENT{Update model}
  \ENDFOR
 \end{algorithmic}
 \end{algorithm*}

\begin{figure}
\centering 
\includegraphics[width=0.45\textwidth]{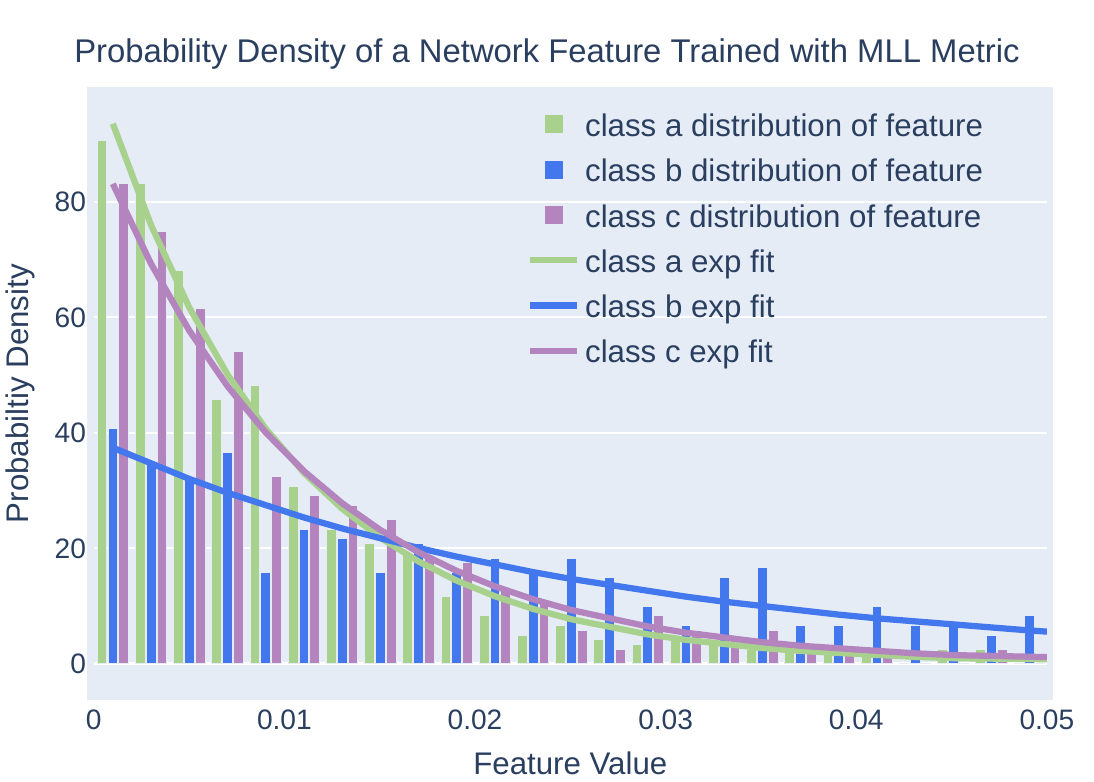}
\caption{Example distributions from the $i$th feature of a network trained with the prototypical paradigm in \cite{ye2020few} and our maximum log likelihood metric. The histograms are the empirical distributions from the same $i$th feature of a random class of miniImageNet validation data. The corresponding lines represent the approximate maximum likelihood estimation of the distribution w.r.t to an exponential fit. This can be compared to the results in Figure \ref{fig:feature_examples} to demonstrate that our metric also results in exponentially distributed features.}
\label{fig:FEAT_MLL_Features}
\vspace{-1em}
\end{figure}

Figure \ref{fig:FEAT_MLL_Features} shows an example of the probability density when training using the prototypical paradigm in \cite{ye2020few} with our maximum log-likelihood metric. This plot is comparable to the plots in Figure \ref{fig:feature_examples}, which  shows that training with our MLL approach results in features that are exponentially distributed.

\subsection{Combining Metrics Using Decision Estimation}
A natural question that arises from the variety of similarity metrics is: {\em what similarity metric produces the best few-shot accuracy}? Unfortunately, there is no well-defined answer to this question. 
The rapidly expanding development of few-shot approaches often makes it challenging to cross-compare methods without standardization, and authors such as Chen et al. and Veilleux et al. have demonstrated that common frameworks often limit reported gains \cite{chen2019closer,veilleux2021realistic}. 
In the past, the authors of Prototypical Networks \cite{snell2017prototypical} argued that a Euclidean metric is superior to a cosine metric; however, the introduction of different network architectures and training paradigms have since achieved better performance with a cosine metric \cite{mangla2020charting, hu2021leveraging}. 

In our experiments section, we carefully isolate the similarity metric to demonstrate that our approach improves the performance over traditional Euclidean and cosine metrics. Further, our MLL approach pushes the conventional prototypical network framework above many state-of-the-art methods that are architecturally more complicated.

We also observed that the network's classification outputs with the three different metrics are highly correlated. As an example, the three networks trained with different metrics were in agreement 73\% of the time on the classifications of miniImageNet's validation data; however, this still leaves 27\% of samples where one or more metric disagrees on a classification. 

\begin{figure*}
     \centering
     \subfloat[Euclidean Scores $\alpha_{euc}(c)$]{\includegraphics[width=.3\textwidth]{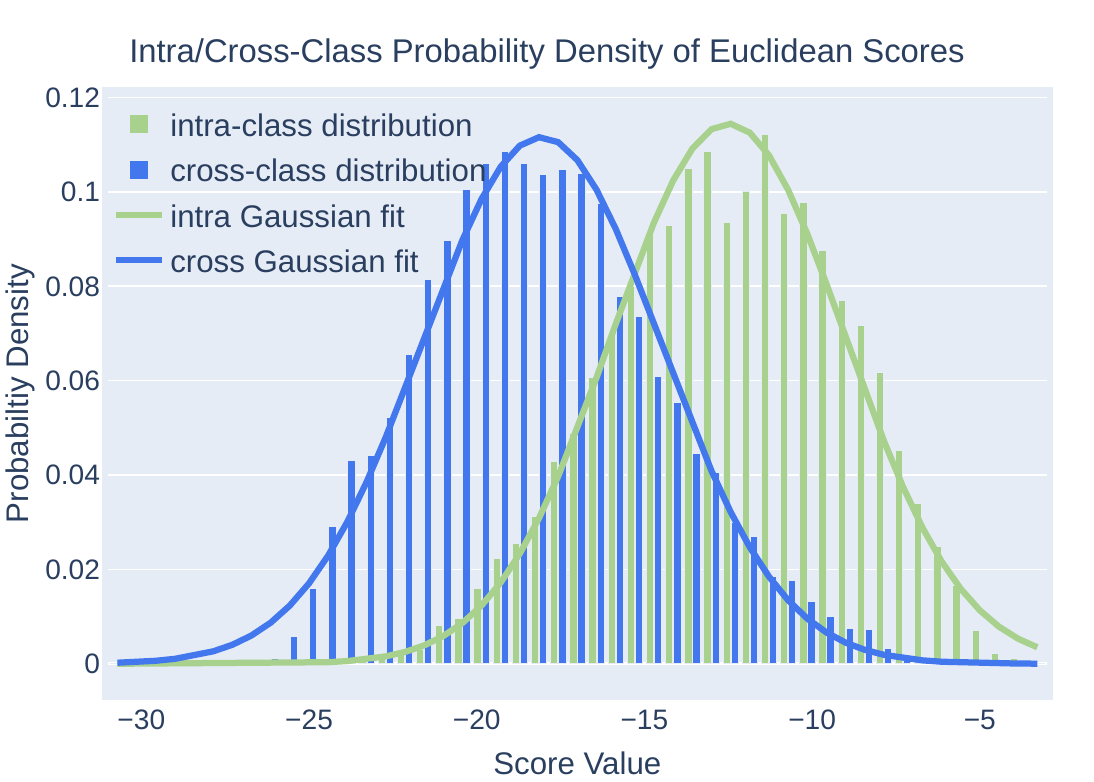}\label{fig:FEAT_Euclidean_Scores}} 
     \subfloat[Cosine scores $\alpha_{cos}(c)$]{\includegraphics[width=.3\textwidth]{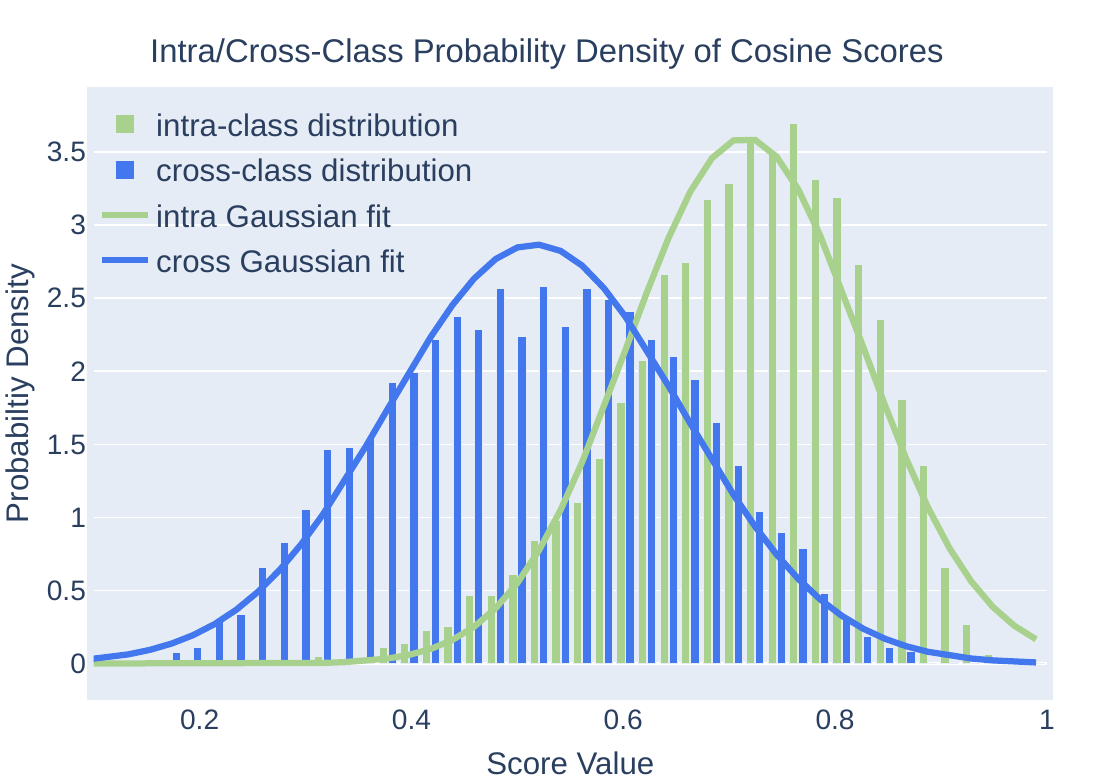}\label{fig:FEAT_Cosine_Scores}} 
     \subfloat[MLL scores $\alpha_{mll}(c)$]{\includegraphics[width=.3\textwidth]{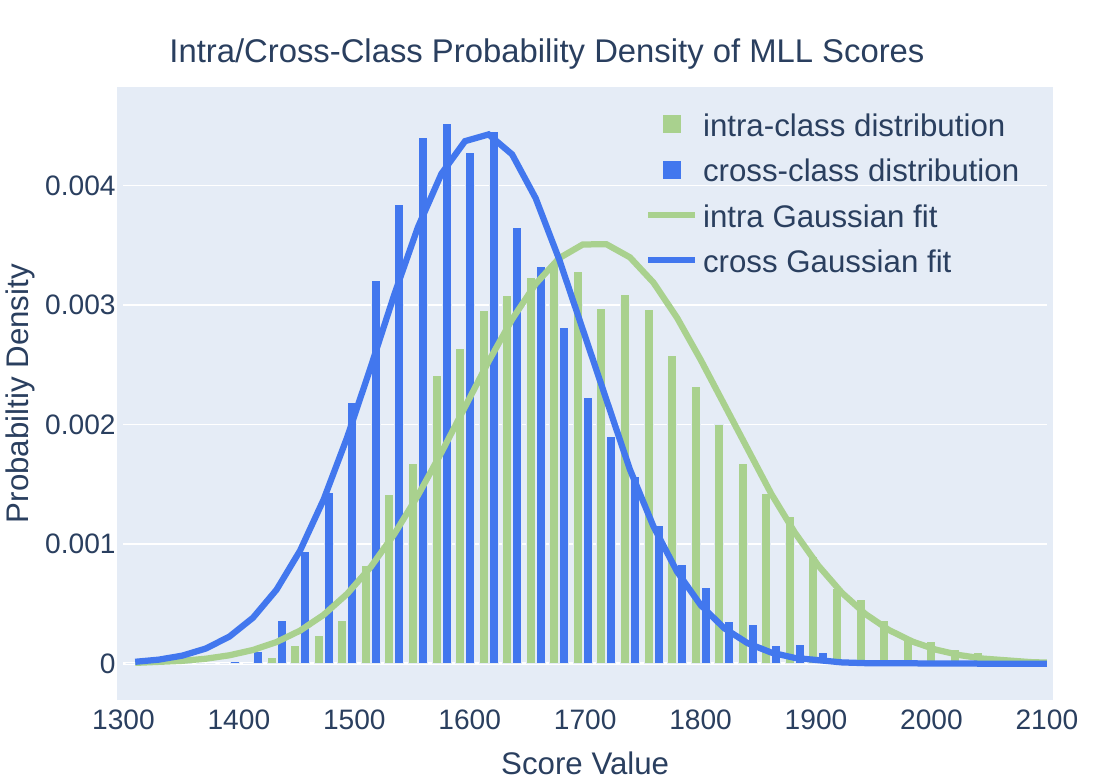}\label{fig:FEAT_MLL_Scores}} 
         \caption{Example distributions from the intra-class and cross-class scores computed with different metrics and their respective trained network. The histograms are the empirical distributions across miniImageNet validation data. The corresponding lines represent the approximate maximum likelihood estimation of the distribution w.r.t to an Gaussian fit. Whereas the network features fit exponential distributions as shown in Figure \ref{fig:feature_examples}, the scores fit well to Gaussian distributions.}
        \label{fig:score_distributions}
        \vspace{-1em}
\end{figure*}

In this section, we demonstrate that the multiple scores can be probabilistically combined to achieve better performance over any single metric. To formulate our approach, we begin by making another fast approximation to the distribution of metric scores.
Specifically, we define three networks trained with Euclidean, cosine, and our maximum log-likelihood approach as $\phi_{euc}$, $\phi_{cos}$, and $\phi_{mll}$, respectively. 
As discussed in the prior section, the corresponding $i$th network feature are represented as ${f}_{\phi_{euc},i}$, ${f}_{\phi_{cos},i}$, and ${f}_{\phi_{mll},i}$. For simplified notation, let the average feature vector over the support set of a network $\phi$ be represented as $\bar{\mathbf{f}}_{\phi} = \frac{1}{|S_c|} \sum_{(\xbf_{sn},y_{sn})\in S_c}\mathbf{f}_{\phi}({\xbf}_{sn})$ and its reciprocal as $\bar{\boldsymbol{\lambda}}_{\phi} = \nicefrac{1}{\bar{\mathbf{f}}_{\phi}}$ where the division is element-wise. Then, the metric scores from each network is given as:
\begin{align}
\alpha_{euc}(c) &= -\left\Vert \mathbf{f}_{\phi_{mll}}(\mathbf{x}_{qm}) - \bar{\mathbf{f}}_{\phi_{euc}} \right\Vert_2^2 \\
\alpha_{cos}(c) &= \frac{\mathbf{f}_{\phi_{cos}}(\mathbf{x}_{qm}) \cdot \bar{\mathbf{f}}_{\phi_{cos}}}{\left\Vert \mathbf{f}_{\phi_{cos}}(\mathbf{x}_{qm})\right\Vert_2^2 \left\Vert \bar{\mathbf{f}}_{\phi_{cos}} \right\Vert_2^2 } \\
\alpha_{mll}(c) &= \sum_{i=1}^I \log\left(\frac{1}{\bar{f}_{\phi_{mll},i}} \right) -  \bar{\boldsymbol{\lambda}}_{\phi_{mll}} \cdot \mathbf{f}_{\phi_{mll}}(\mathbf{x}_{qm})
\end{align}
We isolate the intra-class scores (i.e., $y_{qm} = c$) from the cross-class scores (i.e., $y_{qm} \neq c$) for validation data and plot the probability density of $\alpha_{euc}(c)$, $\alpha_{cos}(c)$, and $\alpha_{mll}(c)$ as shown in Figure \ref{fig:score_distributions}. We let each of these terms form a vector $\boldsymbol{\alpha}(c) = [\alpha_{euc}(c), \alpha_{cos}(c), \alpha_{mll}(c)]^T$.

Although the distributions are not perfectly Gaussian, applying a Gaussian approximation allows us to quickly combine the metrics into an intra-class and cross-class joint probability distribution that take into account the relative correlations through the covariance. 
Further, we show in Section \ref{sec:experiments} that this approximation provides better performance than the state-of-the-art. 
The means ($\boldsymbol{\mu}_{\text{cross}}$, $\boldsymbol{\mu}_{\text{intra}}$) and covariances ($\boldsymbol{\Sigma}_{\text{cross}}$, $\boldsymbol{\Sigma}_{\text{intra}}$) are calculated from the validation scores across the three metrics and the joint cumulative distribution functions are calculated using the Gaussian approximation: 
\begin{align}
    \boldsymbol{\Phi}_{\text{intra}}\left(\boldsymbol{\alpha}(c)\right) &= \int\limits_{-\infty}^{\boldsymbol{\alpha}(c)} \frac{1}{{|\boldsymbol{\Sigma}_{\text{intra}}|^{1/2} (2\pi)^{3/2} }}\text{e}^{ -\frac{1}{2}d_{\text{intra}}^{2}(\alpha)} \,\d\boldsymbol{\alpha} \\
    \boldsymbol{\Phi}_{\text{cross}}\left(\boldsymbol{\alpha}(c)\right) &= \int\limits_{-\infty}^{\boldsymbol{\alpha}(c)} \frac{1}{{|\boldsymbol{\Sigma}_{\text{cross}}|^{1/2} (2\pi)^{3/2} }}\text{e}^{ -\frac{1}{2}d_{\text{cross}}^{2}(\alpha)} \,\d\boldsymbol{\alpha}
\end{align}
where $d_{\text{x}}(\alpha) = (\boldsymbol{\alpha}-\boldsymbol{\mu}_{\text{x}})^\T \boldsymbol{\Sigma}_{\text{x}}^{-1}(\boldsymbol{\alpha}-\boldsymbol{\mu}_{\text{x}})$ is the Mahalanobis distance between the vector $\boldsymbol{\alpha}$ and the Gaussian distribution parameterized by mean $\boldsymbol{\mu}_{\text{x}}$ (the intra or cross class mean) and covariance $\boldsymbol{\Sigma}_{\text{x}}$. Note that the integration is over the three dimensions of $\boldsymbol{\alpha}(c)$. 

Given a query sample $\mathbf{x}_{q}$ with unknown label $y_{q}$, the probability that the sample is a true positive for a given class $c$ is given by the intra-class CDF ($\boldsymbol{\Phi}_{\text{intra}}\left(\boldsymbol{\alpha}(c) \right)$). Similarly, the probability that the sample is a false positive is given by right side tail of the cross-class CDF ($1-\boldsymbol{\Phi}_{\text{cross}}\left(\boldsymbol{\alpha}(c)\right)$). 
Intuitively, the query sample can be represented as a point on a receiver operating characteristic (ROC) plot w.r.t. each class. The class with the highest true positive probability and lowest false positive probability is the most likely class. There is a trade off between the two probabilities and choosing the optimal point is an area of open research \cite{unal2017defining}. We investigated several approaches for the optimal criteria, including the Youden's Index \cite{youden1950index}, the closets to (0,1) criteria, and the Concordance Probability Method \cite{liu2012classification}. We found all three approaches achieved similar results and chose the Youden's Index for the computational simplicity. Formally, the final classification rule is given by: 
\begin{equation}
\widehat{y}_{q} = \argmax_{c\in N} \left[ \boldsymbol{\Phi}_{\text{intra}}\left(\boldsymbol{\alpha}(c) \right) - \left( 1-\boldsymbol{\Phi}_{\text{cross}}\left(\boldsymbol{\alpha}(c)\right) \right) \right]
\label{eq:combined_metric}
\end{equation}

\subsection{Application of MLL metric to Transductive Setting}
The algorithms presented in the previous subsection are inductive algorithms. The combined metric is unique because it {\em{does}} use statistics from the validation data; however, the approach uses the information from only a {\em single} query sample which is consistent with inductive methods. In this section, we consider the application of the exponential MLL metric (presented in Section \ref{sec:MLL_Metric}) for transductive classification where data are evaluated per query batch.

Prototypical networks are based on the principal of estimating the best exemplar embedding for each class ($\bar{\mathbf{f}}_{\phi}$), regardless of the metric used (i.e., Euclidean, cosine, MLL). In the $K$-shot case, there are $K$ known exemplary samples for each class, therefore, the embeddings of exemplary samples are averaged. In the situation where there is a batch of unlabeled samples, $\mathcal{Q}_{u} := \mathcal{Q}_{u1} \cup\, \cdots\, \cup \mathcal{Q}_{uN}$, where the amount of samples is different for each $\mathcal{Q}_{uc}$ (i.e., imbalanced), our objective is to estimate a class embedding that individually incorporates weights for each feature of an unknown sample.

Formally, a prototypical feature, $\bar{\mathbf{f}}_{\phi,c}$, for class $c$ has a corresponding exponential parameter $\bar{\boldsymbol{{\lambda}}}_{c} =  \nicefrac{1}{\bar{\mathbf{f}}_{\phi,c}}$, where the division here is element-wise. 
Thus, the entries of $\bar{\boldsymbol{{\lambda}}}_{c}$ represent the parameter of an exponentially distributed random variable. 
For query feature $\mathbf{f}_{\phi}(\mathbf{x}_{qm})$ the class, $\widehat{y}_{qm}$, can be estimated from equation \eqref{eq:MLL}. 
Intuitively, the exponential distribution has larger densities in the likelihood of the features that are close to zero, especially when $\bar{{\lambda}}_{c,i}$ is large (where $i$ corresponds to the $i$-th feature). 
We observed that larger feature weights helps to create more differences between small and large $\bar{{\lambda}}_{c,i}$ estimates, which subsequently helps classification. This desired feature weighting can be obtained with the cumulative probability between a given $\bar{{\lambda}}_{c,i}$ and $f_{\phi,i}(\mathbf{x}_{qm})$ as:

\begin{equation}
\boldsymbol{\Phi}_{qm}(c) = 1 - \text{e}^{-\boldsymbol{\lambda}_{c} \odot \mathbf{f}(\mathbf{x}_{qm})}
\end{equation}
where $\odot$ is the element-wise Hadamard product. 
The query sample feature embeddings previously estimated to be in class $c$ can be weighted by $\boldsymbol{\Phi}_{qm}(c)$:
\begin{equation}
\bar{\mathbf{g}}_{\phi,c} = \sum_{m: {y}_{qm} = c}\boldsymbol{\Phi}_{qm}(c) \odot \mathbf{f}(\mathbf{x}_{qm})
\end{equation}
where $\bar{\mathbf{g}}_{\phi,c}$ is the weighted average prototype of the query samples.
The new prototypical estimate can then be updated via a simple weighted sum.
\begin{equation}
\bar{\mathbf{f}}_{\phi,c} = (1-\eta)\bar{\mathbf{f}}_{\phi,c} + \eta\bar{\mathbf{g}}_{\phi,c}
\end{equation}

The update process can be repeated for multiple iterations or until a convergence is achieved. The final prototypes, $\bar{\mathbf{f}}_{\phi,c}$, are used to recompute the class, $\widehat{y}_{qm}$, from equation \eqref{eq:MLL}.

Algorithm \ref{alg:MLL_Transductive} presents the pseudo-code using our transductive approach. 
Lines 1-5 achieve the initial inductive estimates of $\widehat{y}_{qm}$. Lines 6 through 11 are the transductive weightings and Lines 12-15 re-estimate $\widehat{y}_{qm}$ for each iteration. Throughout our experiments we use 10 iterations ($I=10$) and $\eta=0.5$. We have found these general hyperparameters work well across datasets and networks and we provide a hyperparameter study in our code repository that demonstrates similar performance for $I \geq 5$ and $0.3 \leq \eta \leq 0.7$.

 \begin{algorithm*}[t] \footnotesize
 \caption{Probability Weighted Maximum Log Likelihood for Transductive Few-Shot pseudo-code}\label{alg:MLL_Transductive}
 \begin{algorithmic}[1]
 \renewcommand{\algorithmicrequire}{\textbf{Input:}}
 \renewcommand{\algorithmicensure}{\textbf{Output:}}
 \REQUIRE{Unlabeled and Imbalanced Query Set $\mathcal{Q}_{u}=\{\mathbf{x}_{q1},\ldots,\mathbf{x}_{qM}\}$, labeled support set, $\mathcal{S}_c=\{(\mathbf{x}_{s1}, {y}_{s1}=c), \ldots, (\mathbf{x}_{sK}, {y}_{sK}=c)\}$, trained few-shot model $\phi$, clip parameter ${\lambda}_{max}$, and hyperparameters for the number of iterations ($I$) and update weight ($\eta$)}
 \ENSURE{class estimates $\widehat{y}_{q_m}$}
 \STATE $\bar{\mathbf{f}}_{\phi,c} = \frac{1}{|\mathcal{S}_c|}  \sum_{(\xbf_{sn},y_{sn})\in S_c}\mathbf{f}_{\phi}({\xbf}_{sn})$ \COMMENT{Compute the average prototype from the support set for each class}
 \STATE $\bar{\boldsymbol{\lambda}}_{c} =  \nicefrac{1}{\bar{\mathbf{f}}_{\phi,c}}$
 \STATE ${\lambda}_{c,i} = \min({\lambda}_{c,i},{\lambda}_{max})$
 \STATE $\alpha_{qm}(c)= \sum_{i}\log({\lambda}_{c,i}) -  \boldsymbol{\lambda}_{c} \cdot \mathbf{f}(\mathbf{x}_{qm})$ \COMMENT{maximum log likelihood, equation \eqref{eq:MLL}}
 \STATE $\widehat{y}_{qm} = \argmax_{c\in N} \left[ \alpha_{qm}(c) \right] $
   \FOR[Each iteration]{$i = 1,\ldots,I$}
   \STATE $\boldsymbol{\Phi}_{qm}(c) = 1 - e^{-\boldsymbol{\lambda}_{c} \cdot \mathbf{f}(\mathbf{x}_{qm})}$  \COMMENT{CDF of given query sample w.r.t. class}
    \FOR[Each each class]{$c = 1,\ldots,N$}
    \STATE $\bar{\mathbf{g}}_{\phi,c} = \sum_{qm: {y}_{qm} = c}\boldsymbol{\Phi}_{qm}(c) \cdot  \mathbf{f}(\mathbf{x}_{qm})$
    \STATE $\bar{\mathbf{f}}_{\phi,c} = (1-\eta)\bar{\mathbf{f}}_{\phi,c} + \eta\bar{\mathbf{g}}_{\phi,c}$
    \ENDFOR
   \STATE $\bar{\lambda}_{c} =  \nicefrac{1}{\bar{\mathbf{f}}_{\phi,c}}$
   \STATE ${\lambda}_{c,i} = \min({\lambda}_{c,i},{\lambda}_{max})$
   \STATE $\alpha_{qm}(c)= \sum_{i}\log({\lambda}_{c,i}) -  \boldsymbol{\lambda}_{c} \cdot \mathbf{f}(\mathbf{x}_{qm})$ \COMMENT{maximum log likelihood, equation \eqref{eq:MLL}}
   \STATE $\widehat{y}_{qm} = \argmax_{c\in N} \left[ \alpha_{qm}(c) \right] $
   \ENDFOR
 \end{algorithmic}
 \end{algorithm*}

\section{Experiments}\label{sec:experiments}
This section demonstrates the performance of our proposed approach for inductive and transductive few-shot learning. In each case, we include the performance of several state-of-the-art methods for comparison. The comparisons are limited to approaches that do not use generative techniques, auxiliary information, or methods that partition images into subsections (e.g., \cite{bendou2022easy,luo2021rectifying}).

\subsection{Datasets and Network Training}
We use the two established benchmark datasets for few-shot classification, namely: miniImageNet and tieredImageNet. Both miniImageNet and tieredImageNet are modified subsets of the ILSVRC-12 dataset designed to have fewer samples per class to emphasize few-shot evaluation. miniImageNet is a balanced set of 600 color images in 100 classes with a standard 64\%/16\%/20\% train/validation/test split. tieredImageNet is 779,165 images partitioned into 608 classes with a 58\%/16\%/26\% (351/97/160 total classes) train/validation/test split. Although tieredImageNet has more classes than miniImageNet, it is designed specifically to separate similar classes in the data split. The authors provide an example of separating ``pipe organ'' and ``electric guitar'' into the training and testing data, respectively, because they both fall under a higher-level classification of ``musical insturment'' in the class hierarchy. In both datasets, the images are resampled to $84\times84$ pixels and the class labels are disjoint between the training, validation, and testing split. 

We follow the same process proposed by Ye et al. \cite{ye2020few} (for ResNet12) and Wang et al. \cite{wang2019simpleshot} (for ResNet18, WRN, and DenseNet121) to train the network architectures. Both approaches use a supervised method that train the classification nodes via cross entropy. Ye et al. then use a transfer learning and episodic training to fine-tune the feature layer w.r.t. a few-shot architecture. Specifically, the fine tuning stage is where the training incorporates the FEAT, Prototypical Euclidean, and our Prototypical MLL approach. ResNet12 is trained for 500 epochs with a batch size of 128 and an initial learning rate of 0.1. The transfer learning aspect of training also has several parameters, including a temperature parameter to adjust the change in loss function and a learning rate set to $10^{-4}$. All the parameters for the supervised and transfer learning of our Prototypical MLL method were kept the same as the Prototypical Euclidean approach to allow for a fair comparison between the different approaches. For ResNet18, WRN, and DenseNet121, we used the pretrained models provided by Veilleux et al. \cite{veilleux2021realistic} which replicated the results of SimpleShot. We used a transfer learning approach, similar to Ye et al., to train each metric (Euclidean, Cosine, and MLL) for an additional 50 epochs.

\subsection{Inductive Few-Shot Performance}
\label{Inductive Few-Shot Performance}
The inductive few-shot performance is a characterization of how well a model classifies a single query sample, $\mathbf{x}_{qm}$, given only the support set, $\mathcal{S}$. $\mathcal{S}$ has $K$, $D$-dimensional samples $\mathbf{x}_i \in \mathbb{R}^D$ with corresponding labels ${y}_i \in \mathcal{Y}$ from $N$ different classes (i.e., $|\mathcal{Y}| = N$). 
In the first experiment we performed an ablation study to directly compare the the Euclidean metric to ours. Specifically, we used the same training procedure, network architecture, hyperparameters, and evaluation as Ye et al. \cite{ye2020few} and Wang et al. \cite{wang2019simpleshot} used to train a prototypical network with Euclidean distance. We isolated a change to the measurement metric from the Euclidean distance to our metric presented in equation \eqref{eq:MLL}, and present the 1-shot and 5-shot results for miniImageNet and tieredImageNet in Table \ref{tab:inductive_Performance_Loss_comparions}. 
Our maximum log-likelihood (MLL) metric consistently outperforms the Euclidean metric. Further, the combined metric consistently achieves the best overall performance. This result that MLL combined with the Euclidean and cosine metrics is one of the key takeaways from this work. The empirical results show that these three metrics combined provide better performance than any one individually and that the MLL metric tends to be best of the three.  

\begin{table}
  \caption{Inductive Few-Shot Performance Comparison of Euclidean and MLL metric}\label{tab:inductive_Performance_Loss_comparions}
  \centering
  \renewcommand{\arraystretch}{1.2}
  \resizebox{\columnwidth}{!}{%
  \begin{tabular}{l c c c>{\columncolor{lightgray}} c>{\columncolor{lightgray}} c }
    \toprule
    \textbf{Model} & \multicolumn{1}{c}{\textbf{dataset}} & \multicolumn{1}{c}{\textbf{backbone}} & \multicolumn{3}{c}{\textbf{1-shot Classification Performance}} \\
    & & & Euclidean & MLL (ours) & Combined (ours) \\
    \toprule
ProtoNet \cite{ye2020few} & miniImageNet	 & ResNet12		& 62.39\%$^*$ & 66.14\% & \textbf{66.28\%}  \\
ProtoNet \cite{wang2019simpleshot} & miniImageNet & WRN & 61.22\%$^*$ & 66.32\% & \textbf{66.59\%}  \\ 
ProtoNet \cite{wang2019simpleshot} & miniImageNet & DenseNet121	& 63.63\%$^*$ & 66.63\% & \textbf{66.88\%}  \\
ProtoNet \cite{ye2020few} & tieredImageNet	 & ResNet12		& 68.23\%$^*$ & 68.54\% & \textbf{70.05\%}  \\
ProtoNet \cite{wang2019simpleshot} & tieredImageNet & WRN & 68.86\%$^*$ & 70.74\% & \textbf{71.77\%}  \\ 
ProtoNet \cite{wang2019simpleshot} & tieredImageNet & DenseNet121	& 69.91\%$^*$ & 71.45\% & \textbf{72.37\%} \\ \toprule
\textbf{} & \multicolumn{1}{c}{\textbf{}} & \multicolumn{1}{c}{\textbf{}} & \multicolumn{3}{c}{\textbf{5-shot Classification Performance}} \\
& & & Euclidean & MLL (ours) & Combined (ours) \\
    \toprule
ProtoNet \cite{ye2020few} & miniImageNet	 & ResNet12		& 80.53\%$^*$ & 81.24\% & \textbf{81.53\%}  \\
ProtoNet \cite{wang2019simpleshot} & miniImageNet & WRN & 81.00\%$^*$ & 81.18\% & \textbf{81.28\%}  \\ 
ProtoNet \cite{wang2019simpleshot} & miniImageNet & DenseNet121	& 82.29\%$^*$ & 81.22\% & \textbf{82.70\%}  \\
ProtoNet \cite{ye2020few} & tieredImageNet	 & ResNet12		& 84.79\%$^*$ & 85.02\% & \textbf{85.23\%}  \\
ProtoNet \cite{wang2019simpleshot} & tieredImageNet & WRN & 85.50\%$^*$ & 85.99\% & \textbf{86.53\%}  \\ 
ProtoNet \cite{wang2019simpleshot} & tieredImageNet & DenseNet121	& 86.42\%$^*$ & 86.62\% & \textbf{86.80\%}  \\ \toprule
  \end{tabular}
  }
$^*$ Euclidean results as reported by the authors.
\vspace{-1em}
\end{table}

\begin{table}
  \caption{Best miniImageNet Inductive Few-Shot Performance}\label{tab:mini_inductive_Performance}
  \centering
  \renewcommand{\arraystretch}{1.2}
  \resizebox{\columnwidth}{!}{%
  \begin{tabular}{l c c c}
    \toprule
    \multirow{3}{2cm}{\textbf{Model}} & \multicolumn{3}{c}{\textbf{5-Way Classification Performance}}\\
    & \multicolumn{1}{c}{\textbf{backbone}} & \multicolumn{1}{c}{\textbf{1-Shot}} & \multicolumn{1}{c}{\textbf{5-Shot}}\\
    \toprule
MetaOptNet$^*$ \cite{lee2019meta} 		& ResNet18	& 62.64\% & 78.63\% \\
SimpleShot$^*$ \cite{wang2019simpleshot}  & DenseNet121	& 61.49\% & 81.48\% \\ 
CTM$^*$ \cite{li2019finding}  & ResNet18 & 64.12\% & 80.51\% \\ 
DeepEMD FCN $^*$ \cite{zhang2020deepemd}  	& ResNet12	& 65.91\% & 82.41\% \\ 
FEAT$^*$ \cite{ye2020few}   & ResNet12 & 66.78\% & 82.05\% \\ 
ProtoNet Euclidean$^*$ \cite{ye2020few}   & ResNet12	& 62.39\% & 80.53\% \\
\rowcolor{lightgray} ProtoNet MLL (ours)  & DenseNet121	& 66.63\% & 81.22\% \\
\rowcolor{lightgray} ProtoNet Combined (ours)  & DenseNet121	& \textbf{66.88\%} & \textbf{82.70\%} \\ \toprule \toprule
  \end{tabular}
  }
 $^*$ Reported as authors' overall best performing results.
\vspace{-1em}
\end{table}

\begin{table}
  \caption{Best tiered ImageNet Inductive Few-Shot Performance}\label{tab:tiered_inductive_Performance}
  \centering
  \renewcommand{\arraystretch}{1.2}
  \resizebox{\columnwidth}{!}{%
  \begin{tabular}{l c c c}
    \toprule
    \multirow{3}{2cm}{\textbf{Model}} & \multicolumn{3}{c}{\textbf{5-Way Classification Performance}}\\
    & \multicolumn{1}{c}{\textbf{backbone}} & \multicolumn{1}{c}{\textbf{1-Shot}} & \multicolumn{1}{c}{\textbf{5-Shot}}\\
    \toprule
MetaOptNet$^*$ \cite{lee2019meta} 		& ResNet18	& 65.99\% & 81.56\% \\
SimpleShot$^*$ \cite{wang2019simpleshot}  & DenseNet121	& 69.91\% & 86.42\% \\ 
CTM$^*$ \cite{li2019finding}  & ResNet18 & 68.41\% & 84.28\% \\ 
DeepEMD FCN $^*$ \cite{zhang2020deepemd}  	& ResNet12	& 71.16\% & 86.03\% \\ 
FEAT$^*$ \cite{ye2020few}   & ResNet12 & 70.80\% & 84.79\% \\ 
ProtoNet Euclidean$^*$ \cite{ye2020few}  	 & ResNet12	& 68.23\% & 84.03\% \\
\rowcolor{lightgray} ProtoNet MLL (ours)  & DenseNet121	& 71.45\% & 86.62\% \\
\rowcolor{lightgray} ProtoNet Combined (ours)  & Densenet121	& \textbf{72.37\%} & \textbf{86.80\%} \\ \toprule \toprule
  \end{tabular}
  }
  $^*$ Reported as authors' overall best performing results.
  \vspace{-1em}
\end{table}

In addition to the ablation study, we compare MLL's overall performance results to state-of-the-art in Tables \ref{tab:mini_inductive_Performance} and \ref{tab:tiered_inductive_Performance} for miniImageNet and tieredImageNet, respectively. Since the absolute best result of each method varies with respect to the network backbone, we report the highest achieving performance of each method + backbone pair. In contrast to the ablation study, this allows a fair comparison between the overall best performance across all methods and backbones. Despite being much simpler than other methods, we observe that our MLL metric achieves high performance across the benchmarks. Further, our combined (MLL+Euclidean+Cosine) method achieves state-of-the-art performance for both datasets. 

\begin{table}
  \caption{miniImageNet Transductive Few-Shot Performance}\label{tab:mini_Transductive_Performance}
  \centering
  \renewcommand{\arraystretch}{1.2}
  \resizebox{\columnwidth}{!}{%
  \begin{tabular}{l c c c }
    \toprule
    \multirow{3}{2cm}{\textbf{Model}} & \multicolumn{3}{c}{\textbf{5-Way Classification Performance}}\\
     & \multicolumn{1}{c}{\textbf{backbone}} & \multicolumn{1}{c}{\textbf{1-Shot}} & \multicolumn{1}{c}{\textbf{5-Shot}}\\
    \toprule
MAML$^*$ \cite{finn2017model-SAMBIB} 	& ResNet18	& 47.6\% & 64.5\% \\
Versa$^*$ \cite{gordon2018versa}  & ResNet18	& 47.8\% & 61.9\% \\ 
Entropy-min$^*$ \cite{dhillon2019baseline}  & ResNet18 & 58.5\% & 74.8\% \\ 
LR+ICI$^*$ \cite{wang2020instance}  	& ResNet18	& 58.7\% & 73.5\% \\ 
PT-MAP$^*$ \cite{hu2021leveraging}   & ResNet18 & 60.1\% & 67.1\% \\ 
LaplacianShot$^*$ \cite{ziko2020laplacian}  	 & ResNet18	& 65.4\% & 81.6\% \\
BD-CSPN$^*$ \cite{liu2020prototype}  	 & ResNet18	& 67.0\% & 80.2\% \\
TIM$^*$ \cite{boudiaf2020information}  	 & ResNet18	& 67.3\% & 79.8\% \\
$\alpha$-TIM$^*$ \cite{veilleux2021realistic}  	 & ResNet18	& 67.4\% & \textbf{82.5\%} \\
\rowcolor{lightgray} Prototypical MLL (ours)  & ResNet18	& \textbf{69.2\%} & 80.7\% \\ \toprule
Entropy-min$^*$ \cite{dhillon2019baseline}  & WRN & 60.4\% & 76.2\% \\ 
PT-MAP$^*$ \cite{hu2021leveraging}   & WRN & 60.6\% & 66.8\% \\ 
SIB$^*$   & WRN & 64.7\% & 72.5\% \\ 
LaplacianShot$^*$ \cite{ziko2020laplacian}  	 & WRN	& 68.1\% & 83.2\% \\
TIM$^*$ \cite{boudiaf2020information}  	 & WRN	& 69.8\% & 81.6\% \\
BD-CSPN$^*$ \cite{liu2020prototype}  	 & WRN	& 70.4\% & 82.3\% \\
$\alpha$-TIM$^*$ \cite{veilleux2021realistic}  	 & WRN	& 69.8\% & \textbf{84.8\%} \\
\rowcolor{lightgray} Prototypical MLL (ours)  & WRN	& \textbf{72.4\%} & 83.3\% \\ \toprule \toprule
$^*$ Results reported by \cite{veilleux2021realistic}
  \end{tabular}
  }\vspace{-1em}
\end{table}

\begin{table}
  \caption{tiered Transductive Few-Shot Performance}\label{tab:tiered_Transductive_Performance}
  \centering
  \renewcommand{\arraystretch}{1.2}
  \resizebox{\columnwidth}{!}{%
  \begin{tabular}{l c c c }
    \toprule
    \multirow{3}{2cm}{\textbf{Model}} & \multicolumn{3}{c}{\textbf{5-Way Classification Performance}}\\
     & \multicolumn{1}{c}{\textbf{backbone}} & \multicolumn{1}{c}{\textbf{1-Shot}} & \multicolumn{1}{c}{\textbf{5-Shot}}\\
    \toprule
Entropy-min$^*$ \cite{dhillon2019baseline}  & ResNet18 & 61.2\% & 75.5\% \\ 
PT-MAP$^*$ \cite{hu2021leveraging}   & ResNet18 & 64.1\% & 70.0\% \\ 
LaplacianShot$^*$ \cite{ziko2020laplacian}  	 & ResNet18	& 72.3\% & 85.7\% \\
BD-CSPN$^*$ \cite{liu2020prototype}  	 & ResNet18	& 74.1\% & 84.8\% \\
TIM$^*$ \cite{boudiaf2020information}  	 & ResNet18	& 74.1\% & 84.1\% \\
LR+ICI$^*$ \cite{wang2020instance}  	& ResNet18	& 74.6\% & 85.1\% \\ 
$\alpha$-TIM$^*$ \cite{veilleux2021realistic}  	 & ResNet18	& 74.4\% & \textbf{86.6\%} \\
\rowcolor{lightgray} Prototypical MLL (ours)  & ResNet18	& \textbf{76.0\%} & 85.4\% \\ \toprule
Entropy-min$^*$ \cite{dhillon2019baseline}  & WRN & 62.9\% & 77.3\% \\ 
PT-MAP$^*$ \cite{hu2021leveraging}   & WRN & 65.1\% & 71.0\% \\ 
LaplacianShot$^*$ \cite{ziko2020laplacian}  	 & WRN	& 73.5\% & 86.8\% \\
BD-CSPN$^*$ \cite{liu2020prototype}  	 & WRN	& 75.4\% & 85.9\% \\
TIM$^*$ \cite{boudiaf2020information}  	 & WRN	& 75.8\% & 85.4\% \\
$\alpha$-TIM$^*$ \cite{veilleux2021realistic}  	 & WRN	& 76.0\% & \textbf{87.8\%} \\
\rowcolor{lightgray} Prototypical MLL (ours)  & WRN	& \textbf{77.8\%} & 86.6\% \\ \toprule \toprule
$^*$ Results reported by \cite{veilleux2021realistic}
  \end{tabular}
  }\vspace{-1em}
\end{table}

\subsection{Class Imbalanced Transductive Few-Shot Performance}
\label{Transductive Few-Shot Performance}
The transductive few-shot performance is similar to the inductive performance with one exception: the classification accuracy is evaluated for a group of samples $\mathcal{Q}_{u}=\{\mathbf{x}_{q1},\ldots,\mathbf{x}_{qM}\}$ together rather than a single query sample, $\mathbf{x}_{qm}$. As discussed in Section \ref{sec:transductive_related_works}, several transductive methods assume that $\mathcal{Q}_{u}$ is formed by sampling an equal number of samples from each class (i.e. balanced). Veilleux et al. demonstrated that the assumption of balanced data is exploited by most transductive algorithms and, subsequently, imbalanced sampling causes significant decreases in classification accuracy. 
The transductive performance  is shown in in Tables \ref{tab:mini_Transductive_Performance} and Table \ref{tab:tiered_Transductive_Performance} for the miniImageNet and tieredImageNet datasets, respectively. We use the same Dirichlet distribution sampling presented by Veilleux et al. \cite{veilleux2021realistic}. That is, $\mathcal{Q}_{u}$ consists of 75 instances sampled with a Dirichlet's concentration parameter of 2. 
The MLL metric is optimized using Algorithm \ref{alg:MLL_Transductive} with $I=10$ and $\eta=0.5$. We observe from the tables that our MLL metric combined with the iterative weighting achieves considerably higher imbalanced classification accuracy for both miniImageNet and tieredImageNet 1-shot performance. Further, the training process for several network backbones requires transductive specific methods. In contrast, our method's network backbone is trained the same way, regardless of the evaluation.

\section{Conclusion}
Despite the many novel architectures, training paradigms, and data conditioning approaches developed in the few-shot literature, the Euclidean distance remains a prominent metric for measuring similarity in the feature space. In this work, we investigated the data distributions of the feature space of common few-shot networks. Motivated by the shape of the distributions, we proposed a fast approximation of a similarity metric based on the maximum log-likelihood of an exponential distribution. With an ablation study, we demonstrated that our proposed metric achieves better classification performance than the Euclidean metric. In addition, we demonstrated that the performance gain holds for different training procedures, models, both miniImageNet and tieredImageNet, and both 1-shot and 5-shot evaluations. Further, we presented a method for combining the Euclidean, cosine, and our score to achieve a combined approach. This combined classification achieves consistently higher performance than our MLL metric and state-of-the-art performance for both miniImageNet and tieredImageNet few-shot performance. Finally, we presented an iterative algorithm to use the MLL metric in a transductive setting. Our method achieves significant performance improvements in 1-shot transductive evaluation when the data is imbalanced and comparable performance in 5-shot despite being a straightforward iterative approach. The code, algorithms, and data are made publicly available through our repository. In that repository we include an additional dataset, CUB, that further reinforces our findings and conclusions.

Our proposed MLL method is specific to the metric (or ``distance'') computation and throughout this work we demonstrated that the prototypical architecture with our metric achieves state-of-the-art performance over more complicated architectures. Several other architectures use Euclidean distance metric in some form. The boarder impact is that our metric has the potential to achieve additional performance gains if incorporated into future architectures. 
Further, our approach achieved performance gains despite the fact that the assumptions of an exponential distribution and the feature independence were a na\"ive to calculate a fast approximation. 
Naturally, this opens the door for future approaches that can achieve performance gains by reducing (or removing) these assumptions. 
\vspace{-1em}

\bibliographystyle{IEEEtran}
\bibliography{MLL_FSL,greg}

\newpage

\vfill

\end{document}